\documentclass[conference]{IEEEtran}
\IEEEoverridecommandlockouts
\usepackage{cite}
\usepackage{amsmath,amssymb,amsfonts}
\usepackage{graphicx}
\usepackage{textcomp}
\usepackage{xcolor}
\usepackage{algorithm, algorithmic}
\usepackage{multirow}
\usepackage{hyperref}

\usepackage{enumitem}
\usepackage{calc}

\def\BibTeX{{\rm B\kern-.05em{\sc i\kern-.025em b}\kern-.08em
    T\kern-.1667em\lower.7ex\hbox{E}\kern-.125emX}}

\newcommand{\bx}{\textbf{x}}
\newcommand{\bv}{\textbf{v}}
\newcommand{\bk}{\textbf{k}}
\newcommand{\bq}{\textbf{q}}

\begin{document}

\title{AutoKG: Efficient Automated Knowledge Graph Generation for Language Models
\thanks{Bohan Chen is supported by the UC-National Lab In-Residence Graduate Fellowship Grant L21GF3606.  This work is supported by NSF grants DMS-2027277 and DMS-2318817.}
\thanks{$^\ast$ Corresponding author (email: bhchenyz@ucla.edu).}
}

\author{\IEEEauthorblockN{Bohan Chen$^\ast$ and Andrea L. Bertozzi}
\IEEEauthorblockA{\textit{Department of Mathematics, University of California, Los Angeles} \\
520 Portola Plaza\\
Los Angeles, CA, 90095, USA}
}


\maketitle

\begin{abstract}
Traditional methods of linking large language models (LLMs) to knowledge bases via the semantic similarity search often fall short of capturing complex relational dynamics. To address these limitations, we introduce AutoKG, a lightweight and efficient approach for automated knowledge graph (KG) construction. 
For a given knowledge base consisting of text blocks, AutoKG first extracts keywords using a LLM and then evaluates the relationship weight between each pair of keywords using graph Laplace learning. 
We employ a hybrid search scheme combining vector similarity and graph-based associations to enrich LLM responses. Preliminary experiments demonstrate that AutoKG offers a more comprehensive and interconnected knowledge retrieval mechanism compared to the semantic similarity search, thereby enhancing the capabilities of LLMs in generating more insightful and relevant outputs.
\end{abstract}


\section{Introduction}
Large Language Models (LLMs) such as BERT \cite{devlin2018bert}, RoBERTA \cite{liu2019roberta}, T5 \cite{raffel2020exploring}, and PaLM \cite{chowdhery2022palm}, are intricately designed architectures equipped with an extensive number of parameters. These models have been rigorously pre-trained on vast and diverse corpora, thereby enabling them to excel in a wide array of Natural Language Processing (NLP) tasks, from language understanding to both conditional and unconditional text generation \cite{tirumala2022memorization, zhou2022learning}. These advancements have been heralded as a step toward higher-bandwidth human-computer interactions. However, their deployment faces significant challenges. On one hand, LLMs exhibit a tendency for 'hallucinations' \cite{welleck2019neural, ji2023survey}, providing plausible yet nonfactual predictions. On the other hand, the black-box nature of LLMs compromises both interpretability and factual accuracy, often resulting in erroneous statements despite memorizing facts during training \cite{petroni2019language, scialom-etal-2022-fine}.

Knowledge in natural language can be externally sourced from a retrievable database, reducing hallucinations and enhancing the interpretability of LLMs \cite{mialon2023augmented}.
Utilizing dense neural retrievers, which employ dense query and document vectors generated by a neural network \cite{asai2021one}, the system can evaluate the semantic similarity to an information-seeking query by calculating the embedding vector similarity across related concepts \cite{lewis2020retrieval, luan2021sparse}.

To go beyond mere semantic similarity in information retrieval and augment the reasoning capabilities of LLMs, two advanced methodologies are particularly transformative: prompt engineering like the Chain-of-thought prompting, and the incorporation of Knowledge Graphs (KGs)\cite{pan2023unifying}. The former, chain-of-thought prompting, provides a framework for advanced reasoning by generating paths of explanations and predictions that are cross-verified through knowledge retrieval \cite{he2022rethinking, trivedi2022interleaving}. While this method offers significant benefits, it is not the primary focus of this study.  As for the latter, KGs offer LLMs a structured and efficient way to address their limitations in factual accuracy and reasoning \cite{xiong2019improving, pan2023unifying}. KGs not only provide accurate and explicit knowledge crucial for various applications \cite{ji2021survey} but are also known for their symbolic reasoning capabilities to produce interpretable results \cite{zhang2021neural}. These graphs are dynamic, continuously evolving with the addition of new knowledge \cite{mitchell2018never}, and can be specialized for domain-specific requirements \cite{abu2021domain}.


In this study, our emphasis is on techniques of automated KG generation and incorporation with LLMs. Most of the works related to these two tasks rely intensively on the ongoing training of neural networks \cite{pan2023unifying, zhong2023comprehensive}, which is both difficult to employ and less flexible for on-the-fly updates. Traditional KG construction approach uses NLP techniques for entity recognition\cite{nadeau2007survey, grishman1996message}, or keyword identification based on term frequency\cite{salton1988term, ramos2003using}, followed by determining relationship strength through word proximity\cite{mintz2009distant}. Current automated techniques necessitate neural network training \cite{luo2023normalizing,wan2021reasoning, wang2023knowledge}. As for the interaction between KGs and LLMs, neural networks are trained to let LLMs understand the information retrieved from KGs \cite{tian2023graph,yasunaga2022deep}. 

The recent advancements in LLMs make us think much more simply about the automatic generation of KGs and the integration of LLMs with KGs. State-of-the-art LLMs such as ChatGPT\footnote{\url{https://openai.com/blog/chatgpt}}, BARD\footnote{\url{https://blog.google/technology/ai/bard-google-ai-search-updates/}}, and LLAMA\cite{touvron2023llama} have demonstrated impressive reasoning capabilities \cite{bang2023multitask, agarwal2023analysing}. Given sufficient information, they can independently execute effective inference. This observation suggests an opportunity to simplify the KG structure: perhaps the intricate relational patterns found in traditional KGs could be simplified into basic strength indicators of association. Consequently, specific relationships are implicitly conveyed to the model through corpus blocks associated with the KG. In addition, we can provide retrieved keywords and the related corpus directly in the prompt rather than training a network to let LLMs understand the retrieved subgraph structure.

Motivated by these ideas, this study makes the following contributions:
\begin{enumerate}
\item We introduce AutoKG, an innovative method for automated KG generation, based on a knowledge base comprised of text blocks. AutoKG circumvents the need for training or fine-tuning neural networks, employs pretrained LLMs for extracting keywords as nodes, and applies graph Laplace learning to evaluate the edge weights between these keywords. The output is a simplified KG, where edges lack attributes and directionality, possessing only a weight that signifies the relevance between nodes.
\item We present a hybrid search strategy in tandem with prompt engineering, which empowers large LLMs to effectively utilize information from the generated KGs. This approach simultaneously searches for semantically relevant corpora based on embedding vectors and the most pertinent adjacent information within the knowledge graphs.
\end{enumerate}

The KG constructed here is a simplified version compared to traditional KGs, which are typically composed of relations in the form of triplets. Firstly, nodes in AutoKG are not entities in the usual sense; they are more abstract keywords. These keywords can represent entities, concepts, or any content that serves as a foundation for search. Additionally, instead of directed edges with specific semantic meanings found in traditional KGs, AutoKG utilizes undirected edges with a single weight value. The node keywords are extracted from the knowledge base with the aid of LLMs, while the graph structure is algorithmically derived. Such a KG can be efficiently stored with just a keyword list and a sparse adjacency matrix. 

Section~\ref{sec: KG_generation} explains the detailed process of automated KG generation, while Section~\ref{sec:hybrid_search} describes the hybrid search method. An essential highlight is that our proposed techniques require no neural network training or fine-tuning. 

\section{Automated KG Generation}\label{sec: KG_generation}
In this section, we introduce our proposed approach, AutoKG, for automated KG generation. The training aspects of the LLM are not the focus of this article. We operate under the assumption that the LLM is already pre-trained and is accompanied by a corresponding vector embedding model. Specifically, we have employed OpenAI's \textit{gpt-4} or \textit{gpt-3.5-turbo-16k} as the LLM and the \textit{text-embedding-ada-002} as the embedding model. 

\begin{figure*}[t]
    \centering
    \includegraphics[width=1\textwidth]{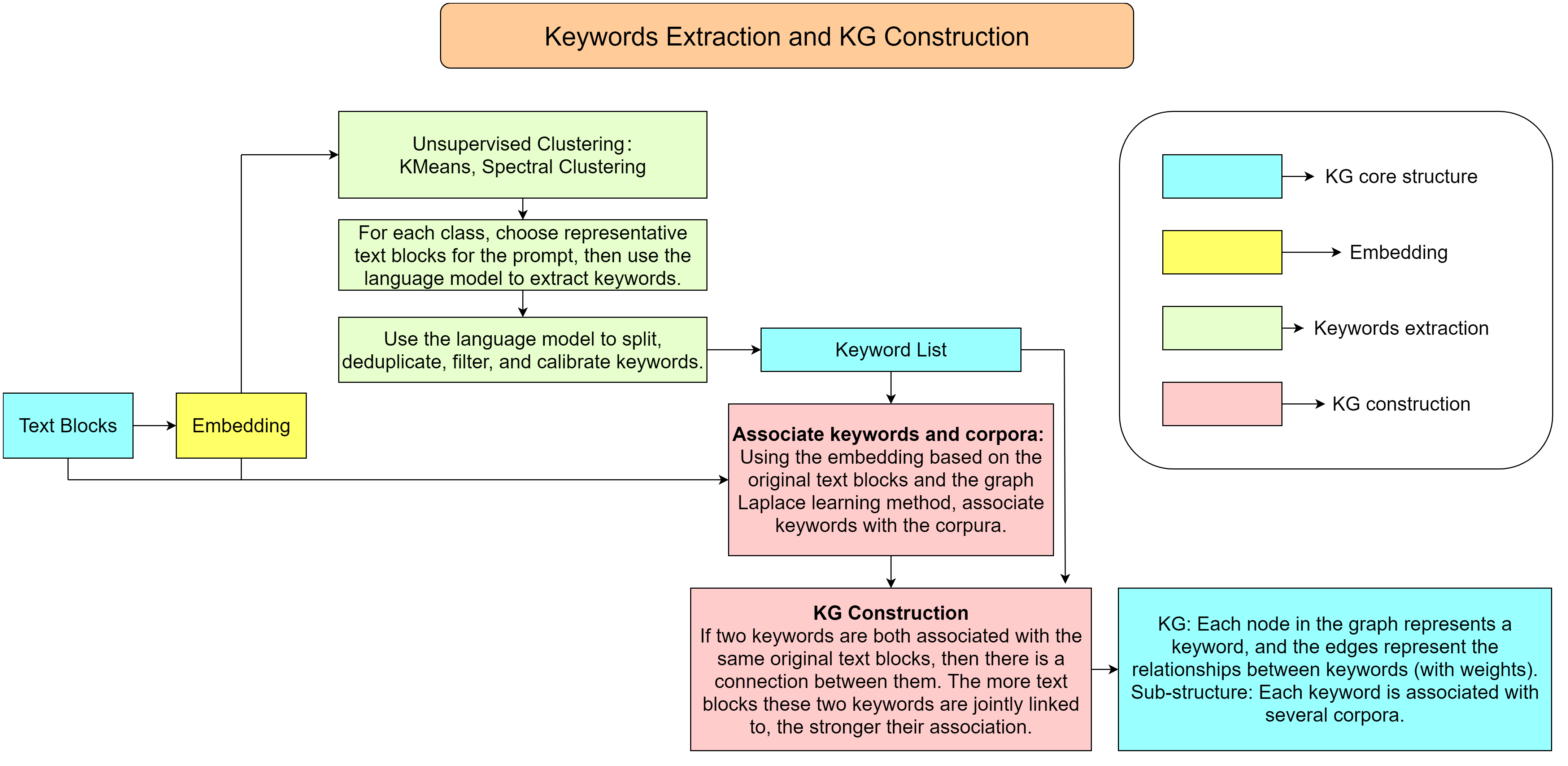}
    \caption{Flowchart of the KG Construction Process. This figure illustrates the different steps involved in the construction of the KG. The blue blocks represent the core components of the KG, yellow blocks indicate the embedding process, green blocks focus on keyword extraction, and the red blocks correspond to the establishment of relationships between keywords and the corpus as well as among the keywords themselves.}
    \label{fig:kg-construction}
\end{figure*}

Consider a scenario involving an external knowledge base, comprised of discrete text blocks. AutoKG constructs a KG where the nodes represent keywords extracted from the external knowledge base. The edges between these nodes carry a single non-negative integer weight, signifying the strength of the association between the connected keywords. AutoKG encompasses two primary steps: the extraction of keywords, which correspond to the nodes in the graph, and the establishment of relationships between these keywords, represented by the edges in the graph. It is worth noting that the pretrained LLM is employed only in the keyword extraction step of the process. Figure~\ref{fig:kg-construction} is the flowchart of the KG construction.

\subsection{Keywords Extraction}\label{sec:keywords_extraction}
Let the external knowledge base be denoted by \(\mathcal{X} = \{\bx_1, \bx_2, \ldots, \bx_N\}\), where each \(\bx_i\) is a block of text with the maximum length of $T$ tokens, represented as a string. The corresponding embedding vectors for these text blocks are encapsulated in \(\mathcal{V} = \{\mathbf{v}(\bx_1), \mathbf{v}(\bx_2), \ldots, \mathbf{v}(\bx_N)\} \subset \mathbb{R}^d\), where $\bv$ is the embedding projection from string to $\mathbb{R}^d$. We extract keywords from the knowledge base \(\mathcal{X}\) with unsupervised clustering algorithms and the assistance of LLMs.  

Algorithm \ref{alg: keywords_extraction} outlines the keyword extraction process. The algorithm takes as input all text blocks and their corresponding embedding vectors \(\mathcal{X}\) and \(\mathcal{V}\), along with pre-defined parameters: \(n\) for the number of clusters, \(c\) for the number of text blocks to select, and \(l_1, l_2\) as keyword extraction parameters. Additionally, the algorithm also utilizes a parameter \(m\) to specify the number of sampled previous keywords. Two unsupervised clustering algorithms, K-means clustering \cite{macqueen1967some, lloyd1982least} and spectral clustering \cite{von2007tutorial}, are applied to cluster the knowledge base. For each cluster identified, we sample \(2c\) text blocks, with \(c\) closest to the cluster center and \(c\) randomly selected, to capture both the global and centered information. The LLM is used twice in this algorithm. First, it extracts keywords from a selection of \(2c\) text blocks, guided by the parameters \(l_1\) and \(l_2\), while avoiding the sampled \(m\) previous keywords. Second, the same LLM is employed to filter and refine the extracted keywords.

The construction of the prompts for these applications strictly follows the format outlined in Table~\ref{tab:prompt_construction}. A specific prompt example for the keyword extraction is given in the Appendix. Specifically, each prompt is formed by concatenating the \textit{Task Information}, \textit{Input Information}, \textit{Additional Requirements}, and \textit{Outputs}. It is essential to note that within each task, the length of the prompt sections corresponding to \textit{Task Information} and \textit{Additional Requirements} is fixed. 

For Task 1, which deals with keyword extraction, the maximum input length is set to \(2cT + m(l_2+1)\), where \(T\) represents the token length of a single text block. Note that each keyword can have a length of up to \(l_2 + 1\) tokens when accounting for potential separators such as commas. Similarly, the maximum output length is \(l_1(l_2 + 1)\), where \(l_1\) is the maximum number of keywords and \(l_2\) is the maximum token length of each keyword. Since Task 1 is applied once for each of the \(n\) clusters generated by the two clustering methods, the total maximum token usage for Task 1 would be \(2n(2cT + (m+l_1)(l_2+1))\). This process yields a maximum of \(2nl_1\) extracted keywords. For Task 2, which involves filtering and refining the keywords, the maximum lengths for both the input and output are governed by the formula \(2n l_1 (l_2 + 1)\). In summary, the maximum usage of tokens $M_\text{tokens KG}$ for the keyword extraction process is 
\begin{equation}\label{eq:max_token_KG}
    M_\text{tokens KG} = 2n(2cT + (m+2l_1)(l_2+1)) + L_F,
\end{equation}
where \(L_F\) is the fixed total length of tokens of the task information and additional requirement parts.

\begin{table*}[t!]
\centering
\caption{Prompt Construction for Different Tasks Using LLM}
\label{tab:prompt_construction}
\begin{tabular}{|l|l|l|l|l|}
\hline
\multirow{2}{*}{\textit{\textbf{ID}}} & \multirow{2}{*}{\textit{\textbf{Task Information}}} & \multirow{2}{*}{\textit{\textbf{Input Information}}} & \multirow{2}{*}{\textit{\textbf{Additional Requirements}}} & \multirow{2}{*}{\textit{\textbf{Outputs}}} \\
& &  &  & \\
\hline
\multirow{3}{*}{1}&\multirow{3}{*}{Keywords Extraction} & \multirow{3}{*}{\begin{tabular}[c]{@{}l@{}}1.Sampled text blocks\\ 2.Sampled previous keywords\end{tabular}} & 1. Avoid previous keywords& \multirow{3}{*}{Extracted Keywords}  \\
& & & 2. Output up to \(l_1\) keywords & \\
&&&3. Each output keyword is at most \(l_2\) tokens&\\
\hline
2&Refining Keywords & Originally extracted keywords & Concentration, deduplication, splitting, and deletion & Refined Keywords \\
\hline
\multirow{3}{*}{3} & \multirow{3}{*}{Response to the Query} & 1. Original query & \multirow{3}{*}{\begin{tabular}[c]{@{}l@{}}Indicate the method used to search for texts and keywords:\\ Direct, via keywords, or KG adjacency search\end{tabular}} & \multirow{3}{*}{Final Response} \\
& & 2. Related text blocks &  &  \\
& & 3. Related keywords &  &  \\
\hline
\end{tabular}
\end{table*}

\begin{algorithm}[h]
\caption{Algorithm for Keyword Extraction in AutoKG}\label{alg: keywords_extraction}
  \begin{algorithmic}[1]
    \REQUIRE All text blocks and their corresponding embedding vectors \(\mathcal{X}\) and \(\mathcal{V}\), pre-defined parameters \(n\) (number of clusters), \(c\) (number of text blocks to select), \(l_1, l_2\) (keyword extraction parameters), \(m\) (number of sampled previous keywords)
    \ENSURE A set of extracted keywords \(\mathcal{K} = \{\bk_1,\bk_2,\ldots,\bk_M\}\)
    \STATE \(\mathcal{K} = \emptyset\)
    \FOR{each clustering algorithm \(P\) in \{k-means, spectral clustering\}}
      \STATE Cluster \(\mathcal{V}\) into \(n\) clusters \(\mathcal{V}^P_i,~i=1,2,\ldots,n\) using algorithm \(P\)
      \FOR{i = 1,2,\ldots,n}
        \STATE Randomly select \(c\) text blocks and \(c\) nearest to the cluster center from cluster \(\mathcal{V}^P_i\)
        \IF{\(|\mathcal{K}| > m\)}
          \STATE Select a subset \(\mathcal{K}_s \subset \mathcal{K}\) such that \(|\mathcal{K}_s| = m\)
        \ELSE
          \STATE \(\mathcal{K}_s = \mathcal{K}\)
        \ENDIF        
        \STATE Include these \(2c\) text blocks and previous keywords \(\mathcal{K}_s\) in a prompt for keyword extraction
        \STATE Use LLM, extract up to \(l_1\) keywords of maximum token length \(l_2\), collected as \(\mathcal{K}^P_i\)
        \STATE Update \(\mathcal{K} = \mathcal{K} \cup \mathcal{K}^P_i\) 
      \ENDFOR
    \ENDFOR
    \STATE Filter and refine \(\mathcal{K}\) using a LLM to obtain the final keyword list
    \RETURN \(\mathcal{K}\)
\end{algorithmic}
\end{algorithm}

\subsection{Graph Structure Construction}
In this section, we detail how to construct a KG based on the keywords extracted in Section \ref{sec:keywords_extraction}. Specifically, we establish whether there are edges between keywords and how to weight these edges. We propose a method based on label propagation on the graph, a step that does not require the involvement of any LLM.

Firstly, we create a graph \(G^t=(\mathcal{X}, W^t)\) where \(\mathcal{X}\) is the set of text blocks serving as the nodes of graph \(G^t\), and \(W^t\) is the weight matrix for the edges. \(W^t_{ij}\) is determined by the similarity between the corresponding embedding vectors \(\bv_i\) and \(\bv_j\). Define the similarity function:
\begin{equation}
    w(\bv_i,\bv_j) = \exp\left(-\frac{\angle(\bv_i,\bv_j)^2}{\sqrt{\tau_i\tau_j}}\right),
\end{equation}
where \(\angle(\bv_i,\bv_j)= \arccos\left(\frac{\bv_i^\top \bv_j}{\|\bv_i\|\|\bv_j\|}\right)\) is the angle between feature vectors \(\bv_i\) and \(\bv_j\). The normalization constant \(\tau_i\) is chosen according to the similarity to the \(K^{th}\) nearest neighbor of \(i\) (i.e., \(\tau_i = \angle(\bv_i, \bv_{i_K})\), where \(\bv_{i_K}\) is the \(K^{th}\) nearest neighbor to \(\bv_i\)).

For computational efficiency, we construct a sparse weight matrix \(W^t\) by considering only the \(K\)-nearest neighbors \cite{arya1998optimal} for each vertex. Let \(x_{i_k},\ k=1,2,\ldots,K\) be the \(K\)-nearest neighbors (KNN) of \(x_i\) (including \(x_i\) itself) according to angle similarity. Define a sparse weight matrix by
\begin{equation}
    \bar{W}^t_{ij} = 
    \begin{cases}
    &w(\bv_i,\bv_j), \ j = i_1,i_2,\ldots,i_K,\\
    &0, \ \text{otherwise}.
    \end{cases}
\end{equation}
\(K\) is chosen to ensure that the corresponding graph \(G^t\) is connected, empirically \(K=30\). We symmetrize the sparse weight matrix to obtain our final weight matrix \(W^t\) by redefining \(W^t_{ij} := (\bar{W}^t_{ij} + \bar{W}^t_{ji})/2\). Note that \(W^t\) is sparse, symmetric, and non-negative (i.e. \(W^t_{ij} \ge 0\)).

Next, we utilize the graph \( G^t=(\mathcal{X}, W^t) \), constructed on text blocks, to establish a keyword KG \( G^k=(\mathcal{K}, W^k) \). Here, \( \mathcal{K} \) is the set of keywords, and \( W^k \) is the weight matrix for the edges. In this matrix, \( W^k_{ij} \) quantifies the strength of association between keywords \( \bk_i \) and \( \bk_j \). Importantly, this association is not semantic but is reflected across the entire corpus in the knowledge base. Specifically, \( W^k_{ij} \) corresponds to the count of text blocks that are simultaneously associated with both keywords \( \bk_i \) and \( \bk_j \).

\begin{algorithm}
\caption{Identifying Keyword to Text Block Association}
\label{alg:keyword_text_relation}
\begin{algorithmic}[1]
\REQUIRE Keyword \( \bk \), Set of text blocks \( \mathcal{X} \), Forward relation parameter \( n_1 \), Backward relation parameter \( n_2 \)
\ENSURE \( \mathcal{X}^\bk \subset \mathcal{X} \), the subset of \( \mathcal{X} \) associated with \( \bk \)
\STATE Obtain the embedding vector \( \bv(\bk) \).
\STATE Find the \( n_1 \) nearest vectors in \( \mathcal{X} \) to \( \bv(\bk) \) (label them as 1) and \( n_2 \) farthest vectors (label them as 0).
\STATE In the text-block graph \( G^t=(\mathcal{X}, W^t) \), use the graph Laplace learning algorithm \cite{zhu2023semi} to label the remaining nodes based on these \( n_1 + n_2 \) labeled nodes. Obtain a real-valued function \( u: \mathcal{X} \rightarrow [0, 1] \) on the graph nodes.
\STATE Define \( \mathcal{X}^\bk = \{\bx_i \in \mathcal{X}: u(\bx_i) \geq 0.5\} \)
\RETURN \( \mathcal{X}^\bk \)
\end{algorithmic}
\end{algorithm}

Algorithm~\ref{alg:keyword_text_relation} establishes the relationship between a keyword and text blocks. The core idea is to select a subset of text blocks that are closest to the keyword as positive data, and another subset that is farthest as negative data. We then employ graph Laplace learning \cite{zhu2023semi} based on the graph structure \( G^t=(\mathcal{X}, W^t) \) that we have previously constructed for text blocks. The graph Laplace learning is a semi-supervised learning method on graphs, utilizing the harmonic property of the solution function \( u: \mathcal{X} \rightarrow [0, 1] \) to diffuse the label values from a subset of labeled nodes to other unlabeled nodes in the graph. The text blocks that are classified towards the positive side (with a node function value \( u \geq 0.5 \)) are considered to be associated with the keyword. 

The association weight \( W^k_{ij} \) between \( \bk_i \) and \( \bk_j \) is defined as follows:
\begin{equation}
    W^k_{ij} = W^k_{ji} = |\mathcal{X}^{\bk_i} \cap \mathcal{X}^{\bk_j}|.
\end{equation}
With this, we complete the construction of the keyword-based KG \( G^k \), which is built upon the text block graph \( G^t \).

\subsection{Time Complexity Analyzation}\label{sec: time_complexity}
This section analyzes the efficiency of the AutoKG method. The token consumption required for KG construction in the AutoKG method has the upper bound according to Eq.~\ref{eq:max_token_KG}. The efficiency of the algorithm is mainly influenced by three aspects:

\begin{enumerate}
    \item \textbf{Constructing the similarity graph based on text blocks} \(G^t=(\mathcal{X}, W^t)\): An approximate nearest neighbor search\cite{arya1998optimal} is employed for KNN search, leading to a complexity of \(O(N\log N)\).

    \item \textbf{Clustering algorithm}: Since both K-means clustering \cite{macqueen1967some, lloyd1982least} and spectral clustering \cite{von2007tutorial} are NP-hard, we bound the complexity by \(I_\text{max}\), the preset maximum number of iterations. Spectral clustering is essentially the Kmeans method augmented with an eigen-decomposition of the graph Laplacian. The time complexity here is mainly dominated by the Kmeans method and is \(O(NndI_\text{max})\), where \(n\) is the number of clusters, and \(d\) is the vector dimension (1536 for OpenAI's embedding model).

    \item \textbf{Graph Laplace learning}: Given that our graph Laplacian matrix is sparse, employing the conjugate gradient method to solve the graph Laplace learning problem results in a time complexity of \(O(\hat{N}\sqrt{\kappa})\), where \(\hat{N}\) represents the count of non-zero elements in the graph Laplacian matrix, and \(\sqrt{\kappa}\) denotes the condition number. We have the upper bound for \(\hat{N}\) as $2KN$, where $K$ is the number of nearest neighbors.
\end{enumerate}

Considering these factors, for large \(N\) and if preconditioning techniques can keep the condition number of the graph Laplacian matrix small, our automated KG construction algorithm should operate with a time complexity of \[O(N\log N + NndI_\text{max} + 2KN\sqrt{\kappa}) = O(N\log N + Nn),\] where the number of clusters practically depends on $N$.

\subsection{Remarks}\label{sec:remarks}
In the process of generating the entire KG, there are several points to be considered:

\begin{itemize}
    \item Although the keywords are extracted from clusters of text blocks, we do not take into account the previous clustering results when establishing the relationship between keywords and text blocks. This is because the same keyword may be included in multiple clusters.
    
    \item When constructing the relationship between keywords, we did not incorporate the embedding vectors of the keywords into the graph for the graph Laplace learning process. There are two reasons for this decision: first, we do not need to update the graph structure when selecting different keywords; second, empirically speaking, the embedding vectors of the keywords tend to be quite distant from the embedding vectors of the text blocks. Therefore, including them in the initial label data for Laplace learning might be meaningless.
\end{itemize}

Our approach considerably outperforms these conventional methods in both keyword extraction and relationship construction. The primary shortcoming of traditional techniques is their reliance on a fixed set of words, leading to a significant loss of related information and often producing overly localized insights. In terms of keyword extraction, our method leverages the capabilities of LLMs, allowing for the refining of keywords that are more central to the topic at hand, rather than merely being high-frequency terms. When it comes to relationship construction, our strategy is grounded in a macroscopic algorithm on graphs of all text blocks. This approach encompasses the information from the entire knowledge base of text blocks, providing a more comprehensive perspective compared to relationships derived from local distances.

\section{hybrid search: Incorporating KG and LLM}\label{sec:hybrid_search}

In this section, we propose a hybrid search approach, based on the KG generated according to Section~\ref{sec: KG_generation}. For a given query, the search results using this hybrid search strategy include not only the text blocks that are semantically related to the query but also additional associative information sourced from the KG. This supplementary data serves to provide more detailed and in-depth reasoning for further analysis by the model. The incorporation of a KG allows us to capture complex relationships between different entities, thereby enriching the contextual understanding of the query.

\begin{algorithm}
\caption{hybrid search Algorithm}
\label{alg:hybrid_search}
\begin{algorithmic}[1]
\REQUIRE Query \( \bq \), embedding vector \( \bv(\bq) \), Parameters \( (s^t_0, s^k_1, s^t_1, s^k_2, s^t_2) \)
\ENSURE Set \( \mathcal{X}_{\text{final}} \) containing text blocks related to \( \bq \), and Set \( \mathcal{K}_{\text{final}} \) containing keywords related to \( \bq \)

\STATE \textbf{Step 1: Vector Similarity Search}
\STATE Find the closest \( s^t_0 \) text blocks in \( \mathcal{X} \) to \( \bv(\bq) \)
\STATE \( \mathcal{X}_0 \leftarrow \) set of closest \( s^t_0 \) text blocks

\STATE \textbf{Step 2: Similar Keyword Search}
\STATE Find the closest \( s^k_1 \) keywords in \(\mathcal{K}\) to \( \bv(\bq) \)
\STATE \( \mathcal{K}_1 \leftarrow \) set of closest \( s^k_1 \) keywords
\STATE For each \( \bk \) in \( \mathcal{K}_1 \), find the closest \( s^t_1 \) text blocks in \( \mathcal{X} \)
\STATE \( \mathcal{X}_1 \leftarrow \) merged set of closest \( s^t_1 \) text blocks for each \( k \) in \( \mathcal{K}_1 \)

\STATE \textbf{Step 3: Keyword Adjacency Search}
\STATE For each \( \bk \) in \( \mathcal{K}_1 \), find \( s^k_2 \) strongest connected keywords according to \( W^k \)
\STATE \( \mathcal{K}_2 \leftarrow \) merged set of \( s^k_2 \) strongest connected keywords for each \( k \) in \( \mathcal{K}_1 \)
\STATE For each \( \bk \) in \( \mathcal{K}_2 \), find the closest \( s^t_2 \) text blocks in \( \mathcal{X} \)
\STATE \( \mathcal{X}_2 \leftarrow \) merged set of closest \( s^t_2 \) text blocks for each \( k \) in \( \mathcal{K}_2 \)

\STATE \( \mathcal{X}_{\text{final}} \leftarrow \mathcal{X}_0 \cup \mathcal{X}_1 \cup \mathcal{X}_2 \)
\STATE \( \mathcal{K}_{\text{final}} \leftarrow \mathcal{K}_1 \cup \mathcal{K}_2 \)
\RETURN \( \mathcal{X}_{\text{final}}, \mathcal{K}_{\text{final}} \)

\end{algorithmic}
\end{algorithm}

In our proposed hybrid search approach, we have devised a multi-stage search process that incorporates both direct text block search as well as keyword-based searching guided by the KG. This process is detailed in Algorithm~\ref{alg:hybrid_search}. Initially, we perform the initial search by computing the text blocks that are closest to the given query embedding vector. Then, we turn to the KG and identify the keywords that are closest to the query, along with text blocks associated with these keywords. Lastly, we identify additional keywords that have the strongest association with the previously identified ones, based on the weight matrix in the KG, and accordingly search for related text blocks. The algorithm returns not just a set of text blocks that are highly relevant to the query, but also a set of keywords that are closely connected to the query. 

To estimate the maximum number of tokens returned by the hybrid search, we consider the maximum number of tokens \( T \) for a single text block and \( l_2 \) for a single keyword. The total number of keywords retrieved will be \( s^k_1 + s^k_1 \cdot s^k_2 \), and the total number of text blocks will be \( s^t_0 + s^k_1 \cdot s^t_1 + s^k_1 \cdot s^k_2 \cdot s^t_2 \). Therefore, the maximum number of tokens $M_\text{tokens QA}$ can be calculated as:
\begin{equation}
    M_\text{tokens QA} = s^k_1 \cdot l_2 \cdot (1 + s^k_2) + T \cdot (s^t_0 + s^k_1 \cdot s^t_1 + s^k_1 \cdot s^k_2 \cdot s^t_2).
\end{equation}

In practical applications, the actual number of tokens obtained through the search often falls below the theoretical maximum. This is because there is substantial overlap between the text blocks and keywords discovered via different search methods. Subsequently, the retrieved information is incorporated into the prompt to enhance the LLM's response to the original query. For details on prompt construction, one may refer to Task 3 in Table~\ref{tab:prompt_construction}. A specific prompt example is provided in the Appendix. Importantly, an adaptive approach can be employed during the prompt construction to ensure that the maximum token limit for the LLM is not exceeded. Text blocks can be added sequentially until the token limit is reached.

\section{Experiments and Results}

In this section, our primary goal is to demonstrate through experiments that our proposed AutoKG approach 
provides significantly better responses while maintaining a comparable efficiency, compared with the retrieval-augmented generation (RAG) method based on semantic vector similarity\cite{lewis2020retrieval, luan2021sparse}. Our approach that combines AutoKG and hybrid search extracts more valuable information for the model than RAG which relies on semantic vector similarity search. 

Unfortunately, we encountered challenges in identifying a suitable dataset to conduct these experiments. We attempted to utilize the WikiWhy dataset \cite{ho2022wikiwhy}, which is designed to evaluate the reasoning capability of models. The dataset comprises approximately 9,000 entries. Each entry contains a paragraph of content, spanning between 100 to 200 words. Based on this content, every entry provides a "why" question along with its corresponding cause-effect relationship and explanation. When we employ the hybrid search based on AutoKG or the semantic vector similarity search of RAG, we can easily retrieve the content corresponding to the given question and instruct the model to answer based on that content. In both methods, the model's responses are almost identical. Since the 9,000 entries are relatively independent of each other, cross-entry data retrieval provided by our method doesn't significantly contribute to answering the questions. 


As a consequence, we adopt qualitative approaches rather than employing numerical metrics to evaluate the experimental performance of our method. First, we provide a simple example to explain why our AutoKG with hybrid search approach has benefits compared to methods based on semantic vector similarity search. Next, we present a detailed example based on all $40$ references of this article and the associated subgraph from the KG used during the query. Finally, we compare the efficiency of hybrid search and semantic vector similarity search from both theoretical and experimental perspectives.

\subsection{A Simple Example: Why We Need KG?}\label{sec:exp_simple_example}
Consider a simple knowledge base that contains text blocks detailing a day in the life of an individual named Alex, along with related information. The core narrative is that after leaving his home in the morning, Alex goes to Cafe A to buy a coffee and then takes a bus to Company B for work. Interspersed within the knowledge base are numerous pieces of granular information such as conversations Alex had with the barista at the cafe, the coffee order details, dialogues on the bus, as well as conversations at his workplace, and so forth.

The point of interest here is how a model would answer the question: \textit{``Was it raining this morning when Alex left his home?''} under the assumption that there is no direct answer to this question and no content about the weather in the knowledge base. We aim to compare the responses given the support information retrieved using our method versus that retrieved through semantic similarity search. Within the knowledge base, there are two indirect pieces of information hinting at the weather conditions:
\begin{enumerate}
    \item Related to Cafe A: ``Many people were chatting and drinking coffee in the square outside Cafe A.''
    \item Related to Company B: ``The car wash located downstairs of Company B was bustling with business today.''
\end{enumerate}
Both these snippets subtly suggest that it was not raining.

Given that the question is primarily about Alex and the weather, the information retrieved from the knowledge base through semantic similarity vector search would only be about Alex (as there is no direct information about the weather). The search results would primarily outline his movements throughout the day. Even with an increase in search entries, it would mostly retrieve additional miscellaneous details, like his coffee order and dialogues. Unfortunately, these details do not contain any hints to infer the day's weather. 

On the other hand, employing AutoKG with a hybrid search approach yields different results. During the KG generation process, we extract keywords such as Alex, Cafe A, and Company B. With the hybrid search, the initial step uses the input question to retrieve the keyword Alex. Then, the adjacency search identifies Cafe A and Company B as related keywords. Subsequently, text blocks are sought based on these keywords, resulting in the identification of implicit weather-related information. 
This example illustrates the utility of the hybrid search. Semantic similarity alone can lack cross-topic connections. It tends to retrieve many minor details within the scope of a given question. When searching with the KG constructed using the AutoKG method, the breadth and diversity of the retrieved information is enhanced. Moreover, prior work has easily substantiated GPT-4's capability to reason effectively with provided clues \cite{bang2023multitask, agarwal2023analysing}. 

From the dialogue record with GPT-4 in the Appendix, it is evident that GPT-4 can accurately infer that it did not rain today when given clues about today's weather. However, when only provided with information about Alex from the semantic similarity vector search, it cannot make any predictions about today's weather.

\subsection{An Example with Article References}\label{sec:exp_reference_example}
We present a concrete example utilizing content from the 42 references cited in this paper. 
The resulting KG is interactively queried using the hybrid search method outlined above. Both the KG generation and subsequent querying processes were performed using the \textit{gpt-3.5-turbo-16k} model, chosen to minimize cost.
The 40 references, once segmented, comprise 5,261 text blocks, each less than 201 tokens in length. For the keyword extraction process, as per Algorithm~\ref{alg: keywords_extraction}, the parameters are: \( n=15, c=15, l_1=10, l_2=3, m=300 \). For Algorithm~\ref{alg:keyword_text_relation}, we use the parameters \( n_1=5 \) and \( n_2=35 \). The entire KG construction consumes 137,516 tokens, which is less than the theoretical maximum of 181,280 tokens given by Eq.~\ref{eq:max_token_KG}. This calculation of the theoretical maximum does not account for the fixed total length of tokens pertaining to task information and additional requirement parts.

The constructed KG comprises 461 nodes (extracted keywords) with its adjacency matrix containing 40,458 non-zero elements. The node with the highest degree in the graph is connected to 289 neighbors. There are 353 nodes whose degree is less than 92, which is 20\% of the maximum possible degree of 460. The entire process of KG construction took approximately ten minutes. All computations, excluding calls to the OpenAI API, are carried out on a CPU with an Intel i9-9900. Both keyword extraction and KG construction take approximately five minutes each.
\begin{figure*}[ht]
    \centering
    \begin{minipage}{.45\textwidth}
        \centering
        \includegraphics[width=\linewidth]{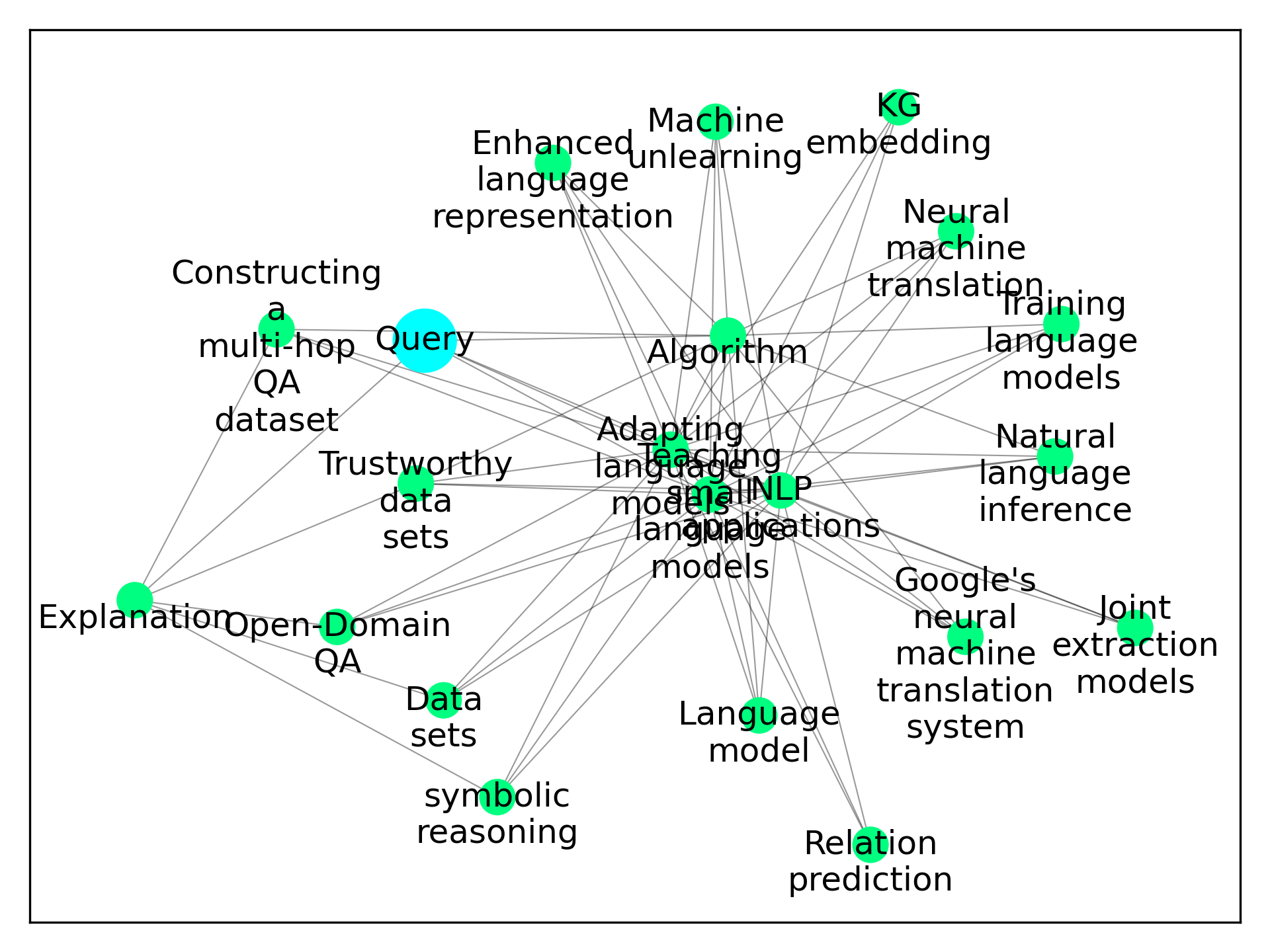}
        \caption{Subgraph Visualization: Keyword Nodes}
        \label{fig:subgraph_vis}
    \end{minipage}%
    \begin{minipage}{.45\textwidth}
        \centering
        \includegraphics[width=\linewidth]{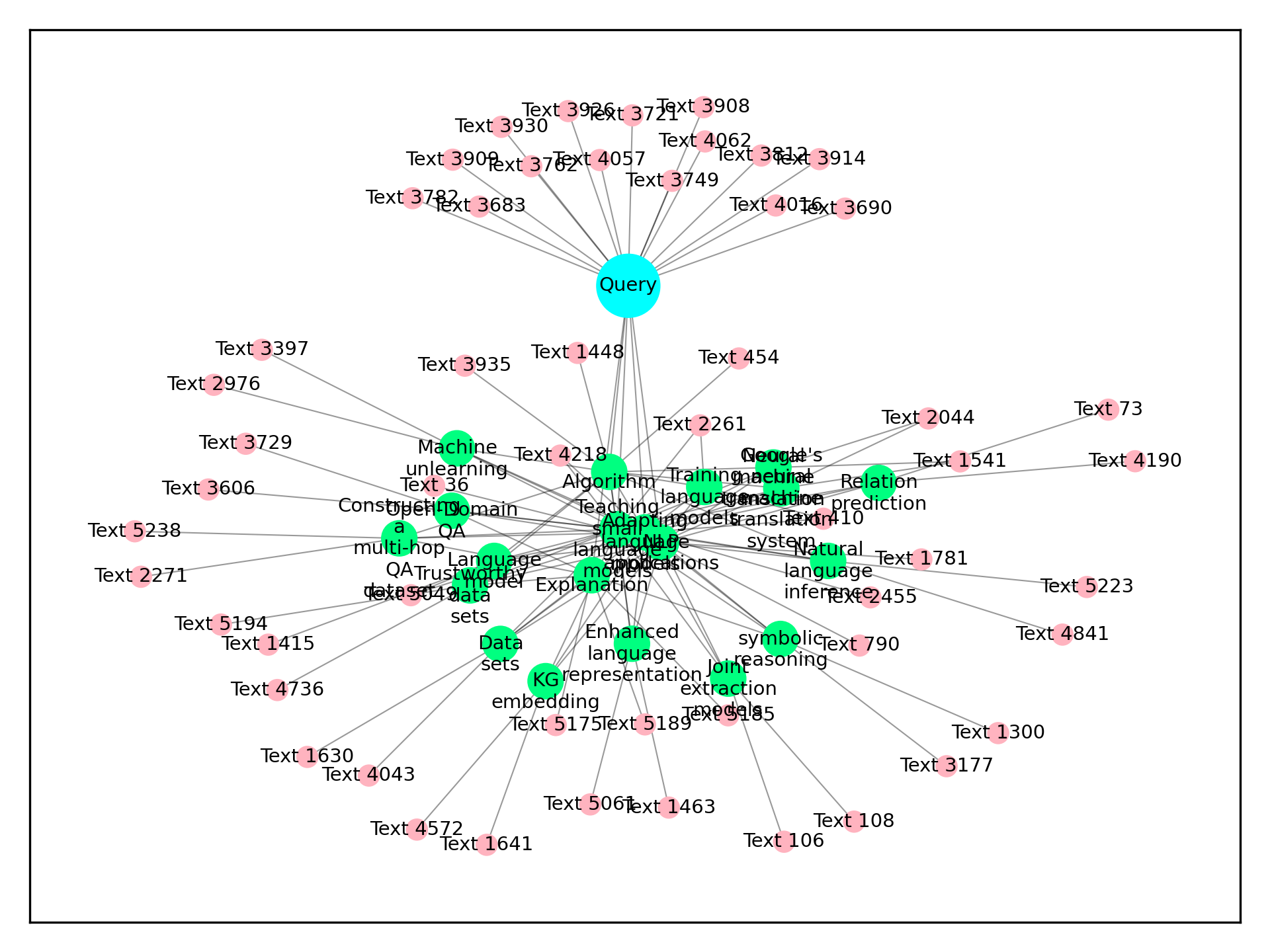}
        \caption{Subgraph Visualization: Keyword and Text Block Nodes}
        \label{fig:subgraph_vis_text}
    \end{minipage}
\end{figure*}
For the subsequent hybrid search described in Algorithm~\ref{alg:hybrid_search}, we use the parameters \( (s^t_0=15, s^k_1=5, s^t_1=3, s^k_2=3, s^t_2=2) \) and ensure, through an adaptive approach, that the input prompt remains under 10,000 tokens in length. The maximum length of response is set as $1024$. As an illustrative example, when querying: \textit{``Please introduce PaLM in detail, and tell me about related applications.''}, the temporary KG structure during the hybrid search is shown in the Figures~\ref{fig:subgraph_vis} and \ref{fig:subgraph_vis_text}. Both images represent subgraphs of the same KG, with the input query depicted in blue nodes. The image on the left (Figure~\ref{fig:subgraph_vis}) showcases only the keyword nodes (in green), while the image on the right (Figure~\ref{fig:subgraph_vis_text}) includes the additionally retrieved text blocks (in pink nodes). The edges displayed are those connecting similar keywords directly retrieved from the query (shown as inner circle nodes in the left figure) as well as edges connecting these similar keywords to the keywords obtained via adjacency search (connecting the inner and outer circles in the left figure). While there may be existing edges between the outer circle keywords, they are omitted from the visualization for clarity.
The model has a lengthy response which is shown in the Appendix. For those interested in further exploration, all pertinent code and test cases are available at \url{https://github.com/wispcarey/AutoKG}. 

\subsection{Efficiency Analyzation}\label{sec:efficiency_analysis}

Given the flexibility in regulating the volume of retrieved information, both the proposed method and the RAG approach can, in theory, support knowledge bases of any size. This means they can encompass any number of text blocks, each subject to the maximum token limit. As outlined in Section~\ref{sec: time_complexity}, the efficiency of the AutoKG method for automated knowledge graph construction is \(O(N\log N)\) when the number of text blocks \(N\) is large.

The constructed keyword KG contains \(M\) keywords where \(M < N\) (empirically, \(M \approx 0.1N\)). During the hybrid search process, with parameters \( (s^t_0, s^k_1, s^t_1, s^k_2, s^t_2) \), the overall time complexity for the search is:
\begin{equation}
O((s^t_0 + s^k_1 \cdot s^t_1 + s^k_1 \cdot s^k_2 \cdot s^t_2)N) + O((s^k_1 + s^k_1\cdot s^k_2)M).
\end{equation}
For the semantic vector similarity search method to retrieve the same volume of text blocks, the time complexity is:
\begin{equation}
O((s^t_0 + s^k_1 \cdot s^t_1 + s^k_1 \cdot s^k_2 \cdot s^t_2)N).
\end{equation}
From the above, it's evident that the time complexity of our hybrid search approach is the same as that of the semantic vector similarity search.
For sufficiently large \(N\), both complexities tend towards \(O(N)\).

Based on the KG generated from the 40 references of this article, as described in Section~\ref{sec:exp_reference_example}, we perform a hybrid search using parameters \( (s^t_0=15, s^k_1=5, s^t_1=3, s^k_2=3, s^t_2=2) \). The theoretical maximum number of text blocks that can be searched using this configuration is 60. For comparison, we conduct a semantic vector similarity search for 30 text blocks. Using a query composed of 50 random characters, we carry out both the hybrid search and semantic vector similarity search methods and record the time taken for each (this includes the embedding computation time). After repeating the experiment 100 times, we calculate the average time taken. The hybrid search method had an average duration of 0.0310 seconds, while the semantic vector similarity search took slightly less, with an average time of 0.0305 seconds. This experiment aligns well with our theoretical analysis of the time complexity.

\section{Conclusion}
This paper addressed the inherent challenges faced by semantic similarity search methods when linking LLMs to knowledge bases. Our method, AutoKG, presents a refined and efficient strategy for automated KG construction. In comparison to traditional KGs, the innovative architecture of AutoKG offers a lightweight and simplified version of KG, shifting the focus from specific entities to more abstract keywords and utilizing weighted undirected edges to represent the associations between keywords. Based on the generated KG, our approach harnesses these capabilities by presenting the LLMs with a more interconnected and comprehensive knowledge retrieval mechanism through the hybrid search strategy. By doing so, we ensure that the model's responses are not only richer in quality but also derive insights from a more diverse set of information nodes.

We tested AutoKG with a hybrid search in experimental evaluations. Because of dataset limitations, our tests were mostly qualitative. The outcome highlights the benefits of our method compared to typical RAG methods with semantic similarity search. In summary, AutoKG provides a valuable step to combine knowledge bases with LLMs. It is computationally lightweight and paves the way for more detailed interactions in LLM applications. Moreover, our hybrid search and the semantic vector similarity search have the same order of time complexity.

Further analysis of the \textit{AutoKG} approach requires the identification or creation of an appropriate dataset to evaluate its integration with LLMs. Wang et al. \cite{wang2023knowledge} developed their own dataset to evaluate a similar idea to ours. While the evaluation criteria should resemble that of RAG, a more structurally intricate and complex dataset is desired. Another avenue for improvement revolves around keyword extraction. Currently, the method leverages prompt engineering; however, future work could explore fine-tuning larger models or even training specialized models to achieve enhanced results.

\section*{Acknowledgment}
The authors acknowledge the assistance of ChatGPT-4 in a first draft of the exposition of the manuscript. 

\bibliographystyle{ieeetr}
\bibliography{ref}

\appendix[Dialogues with LLM]
\subsection*{Dialogue with GPT-4 for Section~\ref{sec:exp_simple_example}}
If we only provide information retrieved from the semantic vector similarity search, GPT-4 can not answer the question.

\noindent\textbf{User:} 

\textit{Answer the question ``Was it raining this morning when Alex left his home?'' with the following information: 1. After leaving his home in the morning, Alex goes to Cafe A to buy a coffee and then takes a bus to Company B for work; 2. Alex bought an iced Americano at Cafe A; 3. Alex talked to the staff at Coffee Shop A about whether they were planning to open a branch; 4. Alex was criticized by his boss at Company B for not completing work on time.
}

\noindent\textbf{GPT-4:}  

\textit{Based on the provided information, there is no direct evidence or indication about the weather condition when Alex left his home in the morning. Therefore, the answer to the question ``Was it raining this morning when Alex left his home?'' is ``The information provided does not specify.''
}

If we provide information about the clue of weather retrieved from the hybrid search based on AutoKG, GPT-4 can complete reasoning and answer the question correctly.

\noindent\textbf{User:} 

\textit{Answer the question ``Was it raining this morning when Alex left his home?'' with the following information: 1. After leaving his home in the morning, Alex goes to Cafe A to buy a coffee and then takes a bus to Company B for work; 2. Many people were chatting and drinking coffee in the square outside Cafe A; 3. The car wash located down-stairs of Company B was bustling with business today.}

\noindent\textbf{GPT-4:} 

\textit{Based on the provided information, it is likely that it was not raining this morning when Alex left his home. The clue is that ``many people gathered in the square in front of the Cafe A, drinking coffee and chatting,'' which indicates a pleasant outdoor environment, not typical of rainy weather. Additionally, the car wash shop doing very well suggests it might be a clear day since car washes tend to be less busy on rainy days.
}

\subsection*{Dialogue with GPT-3.5 for Section~\ref{sec:exp_reference_example}}
\noindent\textbf{Input query:}

\textit{Please introduce PaLM in detail, and tell me about related applications.}

\noindent\textbf{Preprocess:}

With the input query, information is retrieved from the knowledge base through the hybrid search process based on the KG. The prompt is reconstructed with the retrieved information and sent to the \textit{gpt-3.5-turbo-16k} model for the final output.

\noindent\textbf{LLM response:}

\textit{PaLM, which stands for Pathways Language Model, is a large-scale language model that has been developed to achieve state-of-the-art results in few-shot learning across a wide range of natural language processing (NLP) tasks. It is a dense Transformer language model with 540 billion parameters and has been trained on 780 billion tokens of high-quality, diverse text.
}

\textit{The development of PaLM has been driven by the goal of exploring novel architectural choices and training schemes in order to create a large-scale, modularized system with broad generalization capabilities across multiple modalities. PaLM has been designed to be highly scalable, and it leverages the scaling capabilities of the Pathways infrastructure, which enables training a single model across thousands or tens of thousands of accelerator chips in an efficient manner.
}

\textit{PaLM has demonstrated outstanding performance on various NLP tasks. It has achieved breakthrough results on tasks such as open-domain closed-book question answering, cloze and completion tasks, common sense reasoning, in-context reading comprehension, and more. It has also shown strong capabilities in multilingual tasks and source code generation. In fact, PaLM has outperformed the state-of-the-art on a suite of multi-step reasoning tasks and has even surpassed the average human performance on the recently released BIG-bench benchmark.
}

\textit{The primary application of PaLM is research on language models, including NLP applications such as machine translation and question answering. It is being used within Google for research on a variety of open-ended text and code generation tasks, including reasoning and code synthesis. PaLM has also been evaluated for its translation capabilities across a variety of language pairs and settings, particularly focusing on English-centric language pairs.
}

\textit{It is important to note that PaLM is primarily designed for research purposes and has not been extensively tested in settings outside of research. Further analysis is required before using PaLM for downstream applications. Additionally, ethical considerations and risks associated with large language models, including potential misuse, have been discussed in the context of PaLM.
}

\textit{In conclusion, PaLM is a highly scalable language model that has achieved state-of-the-art results in few-shot learning across a wide range of NLP tasks. Its applications include research on language models, machine translation, question answering, code generation, and more. 
}

\appendix[Prompt Examples]
\noindent\textbf{Task: Keyword Extraction}\\
This prompt requires the following preset variables: 1. 'main\_topic', a high-level topic(s) of the knowledge base; 2. '\(l_1\)', the maximum number of extracted keywords; 3. '\(l_2\)', the maximum length of each keyword; 4. 'language', the language of output keywords; 5. 'previous keywords', a list of previously extracted keywords. Once the \textbf{text blocks} are sampled from a certain cluster, we use the following prompt for keyword extraction:\\
\noindent\textbf{Prompt for Keyword Extraction:}\\
\textit{You are an advanced AI assistant, specializing in analyzing various pieces of information and providing precise summaries. Your task is to determine the core theme in the following series of *-separated information fragments, which are delimited by triple backticks. Ensure your answer focuses on the topic and avoids including unrelated content. DO NOT write complete sentences. }
\\\textit{You should obey the following rules when doing this task:
1, Keywords in your answer should related to the topic 'main\_topic';
2, Your answer should include at most '\(l_1\)' keywords;
3, Each keyword should be at most '\(l_2\)' words long;
4, avoid already appeared theme keywords, marked inside \(\langle\rangle\);
5, Write your answer in 'language';
6, Separate your output keywords with commas (,);
7, Don't include any symbols other than keywords.}
\\\\\textit{Information:' ' 'text blocks' ' '}
\\\\\textit{Please avoid the following already appeared theme terms: 
\(\langle\)'previous keywords'\(\rangle\)}
\\\textit{Your response: }

\noindent\textbf{Task: Incorporation between KGs and LLMs}\\
For a given query \( \bq \), we search for its related text blocks \(\mathcal{X}_{\text{final}}\) and keywords \(\mathcal{K}_{\text{final}}\) according to the hybrid search algorithm (Algorithm~\ref{alg:hybrid_search}). Given a preset variable 'language' for the output language, we use the following prompt to provide retrieved information from the KG and original knowledge base:\\
\noindent\textbf{Prompt for Query Response:}\\
\textit{I want you to do a task, deal with a query, or answer a question with some information from a knowledge graph. You will be given a set of keywords directly related to a query, as well as adjacent keywords from the knowledge graph. Relevant texts will be provided, enclosed within triple backticks. These texts contain information pertinent to the query and keywords.}
\\\textit{Please note, you should not invent any information. Stick to the facts provided in the keywords and texts. These additional data are meant to assist you in accurately completing the task. Your response should be written in 'language'.}
\\\textit{Avoid showing any personal information, like Name, Email, WhatsApp, Skype, and Website in your polished response.}
\\\\\textit{Keywords information (directly related to the query or find via the adjacent search of the knowledge graph): \(\mathcal{K}_{\text{final}}\)}
\\\\\textit{Text information: ' ' ' \(\mathcal{X}_{\text{final}}\) ' ' '}
\\\\\textit{Your task: \( \bq \)}
\\\textit{Your response:}

\end{document}